\documentclass[10pt,twocolumn,letterpaper]{article}

\usepackage[pagenumbers]{wacv} 
\definecolor{wacvblue}{rgb}{0.21,0.49,0.74}
\usepackage[pagebackref,breaklinks,colorlinks,allcolors=wacvblue]{hyperref}
\usepackage{import}
\usepackage{dsfont}
\usepackage{url}
\usepackage{multirow} 
\usepackage{graphicx}
\usepackage{amsmath}
\usepackage{amsthm}
\usepackage{amssymb}
\usepackage{booktabs}
\usepackage{ulem}
\usepackage[T1]{fontenc}

\title{GLACIA: Instance–Aware Positional Reasoning for Glacial Lake Segmentation via Multimodal Large Language Model}

\author{
    Lalit Maurya$^{1,*}$ \quad
    Saurabh Kaushik$^{2,*}$\quad
    Beth Tellman$^{2}$\\[0.2cm]
    $^{1}$Portsmouth AI and Data Science Centre (PAIDS), School of Computing, University of Portsmouth, \\Portsmouth, PO1 3HE, UK \\
    $^{2}$Center for Sustainability and the Global Environment (SAGE), University of Wisconsin–Madison, \\Madison, WI, 53726 USA \\
    {\tt\small lalit.maurya@port.ac.uk,
    skaushik8@wisc.edu,
    beth.tellman@wisc.edu}
}

\begin{document}
\maketitle
\let\thefootnote\relax\footnotetext{${}^{*}$These authors contributed equally.}
\begin{abstract}
Glacial lake monitoring bears great significance in mitigating the anticipated risk of Glacial Lake Outburst Floods. However, existing segmentation methods based on convolutional neural networks (CNNs) and Vision Transformers (ViTs), remain constrained to pixel-level predictions, lacking high-level global scene semantics and human-interpretable reasoning. To address this, we introduce GLACIA (\textbf{G}lacial \textbf{LA}ke segmentation with \textbf{C}ontextual \textbf{I}nstance \textbf{A}wareness), the first framework that integrates large language models with segmentation capabilities to produce both accurate segmentation masks and corresponding spatial reasoning outputs. We construct the Glacial Lake Position Reasoning (GLake-Pos) dataset pipeline, which provides diverse, spatially grounded question–answer pairs designed to overcome the lack of instance-aware positional reasoning data in remote sensing. Comparative evaluation demonstrate that GLACIA (mIoU: 87.30) surpasses state-of-the-art method based on CNNs (mIoU: 78.55 – 79.01), ViTs (mIoU: 69.27 – 81.75), Geo-foundation models (mIoU: 76.37 – 87.10), and reasoning based segmentation methods (mIoU: 60.12 – 75.66). Our approach enables intuitive disaster preparedness and informed policy-making in the context of rapidly changing glacial environments by facilitating natural language interaction, thereby supporting more efficient and interpretable decision-making. The code is released on \url{https://github.com/lalitmaurya47/GLACIA}

\end{abstract}
    
\section{Introduction}
\label{sec:intro}

Expansion and development of new glacial lakes are direct results of climate-driven glacier meltwater collected in glacial surficial depressions, and the coalescence of several smaller lakes that gives birth to larger lakes \cite{zhang2024heterogeneous}. The continuous sprawling of such high-mountain glacial lakes could result into sudden discharge of water triggered by overtopping, dam break, anomalous high precipitation, slope failure, tectonic activity, or a combination of these factors \cite{richardson2000overview,veh2020hazard}.  In recent times, we have witnessed an increasing risk of Glacial Lake Outburst Floods (GLOFs) \cite{zheng2021increasing, kaushik2024increasing}, which cause large-scale destruction and fatalities in downstream regions across various mountainous regions \cite{shugar2021massive,sattar2025sikkim}. Thus, continuous monitoring of glacial lakes is of utmost importance for disaster preparedness and mitigating the impact of such anticipated hazards. Most automated glacial lake mapping methods are limited to generating segmentation masks using CNNs \cite{kaushik2022automated,tang2024automatic,hu2024glacial}. Recent developments in computer vision, including multimodal Large Language Models (LLM) \cite{flamingo,llava,kuckreja2024geochat,qwen2vl} allows integration of human-readable positional reasoning, offering potential to advance glacial lake segmentation. However, LLM based approaches remain focused on text generation and struggle with vision tasks that demand fine-grained spatial outputs. This approach leverage high-level reasoning and contextual cues rather than exhaustive pixel-level analysis, providing ample scope for improvement in glacial lake mapping along with text description of lake and scene characteristics.
\begin{figure*}[h]
    \centering
    \includegraphics[width=1\linewidth]{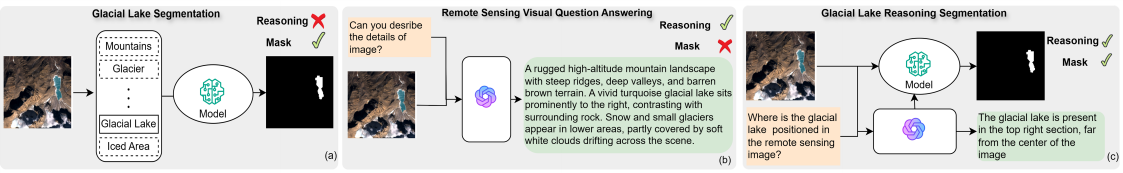}
    \caption{Conceptual shift from traditional segmentation (a) and VQA-based reasoning (b) to our reasoning-driven paradigm (c), which unifies accurate instance-specific masks with interpretable positional reasoning.}
    \label{fig:res_seg_fig}
\end{figure*}

To bridge this gap, we introduce GLACIA, a framework that integrates instance–aware positional reasoning with multispectral features for context-aware segmentation. Unlike conventional methods, our approach leverages a large language model (LLM) to generate reasoning-rich descriptions that capture both the number of lakes and their spatial arrangement within a secene. A hybrid of the Prithvi Geo-Foundation Model (GFM) and a CNN network extracts multispectral features. Fig. \ref{fig:res_seg_fig} highlights the conceptual shift: traditional segmentation (Fig. \ref{fig:res_seg_fig}(a)) lacks reasoning, while visual question answering (VQA) (Fig. \ref{fig:res_seg_fig}(b)) provides descriptive reasoning without actionable segmentation. Our paradigm (Fig. \ref{fig:res_seg_fig}(c)) unifies both, producing interpretable positional reasoning and accurate, instance-specific masks. This design introduces two core capabilities: (1) multimodal reasoning, which integrates LLM-derived lake counts, quadrant cues, and relative positioning (e.g., “near the center,” “top-right”) with visual evidence; and (2) instance-aware spatial localization, which generates distinct segmentation masks for each lake based on reasoning-derived positions. 
GLACIA acts as a human-aligned, intelligent assistant, capable of answering user queries while visually indicating the positions of lakes and glacial mask in an interpretable manner. Our Contributions are:

\begin{itemize}
    \item Creation of the Glacial Lake Position Reasoning (GLake-Pos) dataset pipeline, which generates diverse, spatially grounded question–answer pairs for glacier lake segmentation.
    \item Development of GLACIA, the first position reasoning model that combines multimodal vision–language learning with a Prithvi-Res Encoder for glacial lake mapping.  
    \item Open-sourcing of the dataset pipeline, model, and codebase to support community-driven LLM-assisted remote sensing applications.
    \item Extensive experiments demonstrating GLACIA’s effectiveness in position reasoning and referring segmentation for glacier lake detection and mapping.
\end{itemize}

\section{Related Work}
\label{related_work}
\textbf{Remote Sensing Glacier Lake Segmentation.} Glacial lake segmentation is widely carried out using CNNs with Landsat 8 and Sentinel-2 sensors \cite{tom2025monitoring}. The U-Net and DeepLab variant architectures dominate the literature, with 18 and 6 studies published so far \cite{tom2025monitoring}. Most CNN-based methods have proven effective for clear lake characteristics; however, they exhibit severe limitations in cases of frozen lakes, lake turbidity, small lakes ( $\leq 0.1 \,\text{km}^2$ ), shadows, cloud cover, and misclassification of streams water originating from glacier snout \cite{tom2025monitoring}. Recently, \cite{jiang2025glacial} demonstrated the successful application of a pretrained Prithvi encoder for glacial lake mapping, extending its capability to process multi-sensor data via fusion using a Squeeze-and-Excitation layer.

\noindent\textbf{MultiModal Large Language Model.} Recent advances in LLMs have driven significant progress in unified vision–language reasoning, yet pixel-level interpretation in remote sensing (RS) remains substantially limited. Foundational multimodal LLMs such as Florence-2 \cite{xiao2024florence} showcased instruction-driven unification of text, detection, and segmentation, while LISA \cite{lisa} connected LLMs with segmentation decoders (e.g., SAM \cite{kirillov2023segment}) using the \textbf{[}SEG\textbf{]} token to enable explicit language-guided mask prediction. Subsequent frameworks like GSVA \cite{xia2024gsva} and GLaMM \cite{Rasheed_2024_CVPR} expanded segmentation to multi-target and conversational scenarios, and Text4Seg \cite{lan2024text4seg} reframed segmentation as text generation. These advances motivated the adaptation of LLM-guided segmentation paradigms for RS imagery. 
In RS, early multimodal LLMs primarily addressed image-level understanding. RSGPT \cite{hu2025rsgpt} adapted pretrained MLLMs to tasks such as captioning and image-level VQA. Later models, including GeoChat \cite{kuckreja2024geochat}, EarthGPT \cite{zhang2024earthgpt} and SkyEyeGPT \cite{zhan2025skyeyegpt}, extended these capabilities to region-level grounded interactions, enabling localized captioning, region-specific question answering, and visual grounding tasks. Despite this progression, pixel-level RS MLLMs remain nascent. Models such as RRSIS \cite{yuan2023rrsis} and RMSIN \cite{liu2024rotated} pioneered pixel-level vision–language interfaces but were limited to single-instance referring segmentation, lacking multi-object reasoning, implicit segmentation,  However, all current pixel-level attempts operate exclusively in the visible spectrum, leaving multispectral, hyperspectral, and SAR bands entirely unaddressed in LLM-driven segmentation.
Consequently, robust pixel-level RS MLLMs capable of reasoning across multi-modal spectral data remain an important and unresolved research frontier.

\section{GLake-Pos Dataset Preparation}
Development of various multimodal or description-based datasets—such as RSVQA \cite{lobry2020rsvqa}, RSICD \cite{lu2017exploring}, XLRS-Bench \cite{wang2025xlrs}, \cite{danish2025geobench} and RemoteCLIP \cite{liu2024remoteclip} has gained wide scientific attention, to enable vision-language models (VLMs) to perform image captioning, visual question answering, scene classification, and general geospatial reasoning. Despite all these developments, dataset for precise positional reasoning and instance-level understanding is still lacking, which might have direct implication for many earth observation applications including glacial lake detection. To address this gap, we propose the GLake-Pos dataset pipeline,  to generate a Question Answering dataset designed to include conversations between a end user and an assistant, focusing on lake position, lake numbering, and related environmental information. 

Our dataset pipeline consists of two stages. In the first stage, we prepare question–answer templates using GPT-4, generating 50 templates for training and testing. These templates are validated by domain experts to ensure correctness and contextual relevance, so that when combined with lake identifiers and positional information, the resulting sentences are coherent and informative. In the second stage, we perform instance-aware position extraction and mapping. From each segmentation mask $X_{mask}$, we derive bounding boxes ${\{x, y, w, h\}}$ representing the location of individual lakes and compute their center points $x_{\text{center}} = \left\{ x + \frac{w}{2}, \; y + \frac{h}{2} \right\}$.
The image is divided into four quadrants—top left, top right, bottom left, and bottom right—and each lake is assigned a positional label based on its center. For lakes near the geometric center, a distance threshold is applied to assign a “near center” label. Additionally, each detected lake receives a unique identifier (Lake 1, Lake 2, etc.), enabling instance-aware reasoning that distinguishes multiple lakes within the same scene. Finally, positional information and identifiers are integrated with the validated templates to produce a dataset of spatially grounded glacial lake descriptions. Further implementation details are provided in the supplementary material.

\section{GLACIA}
\label{sec:method}
\begin{figure*}
    \centering
    \includegraphics[width=0.9\linewidth]{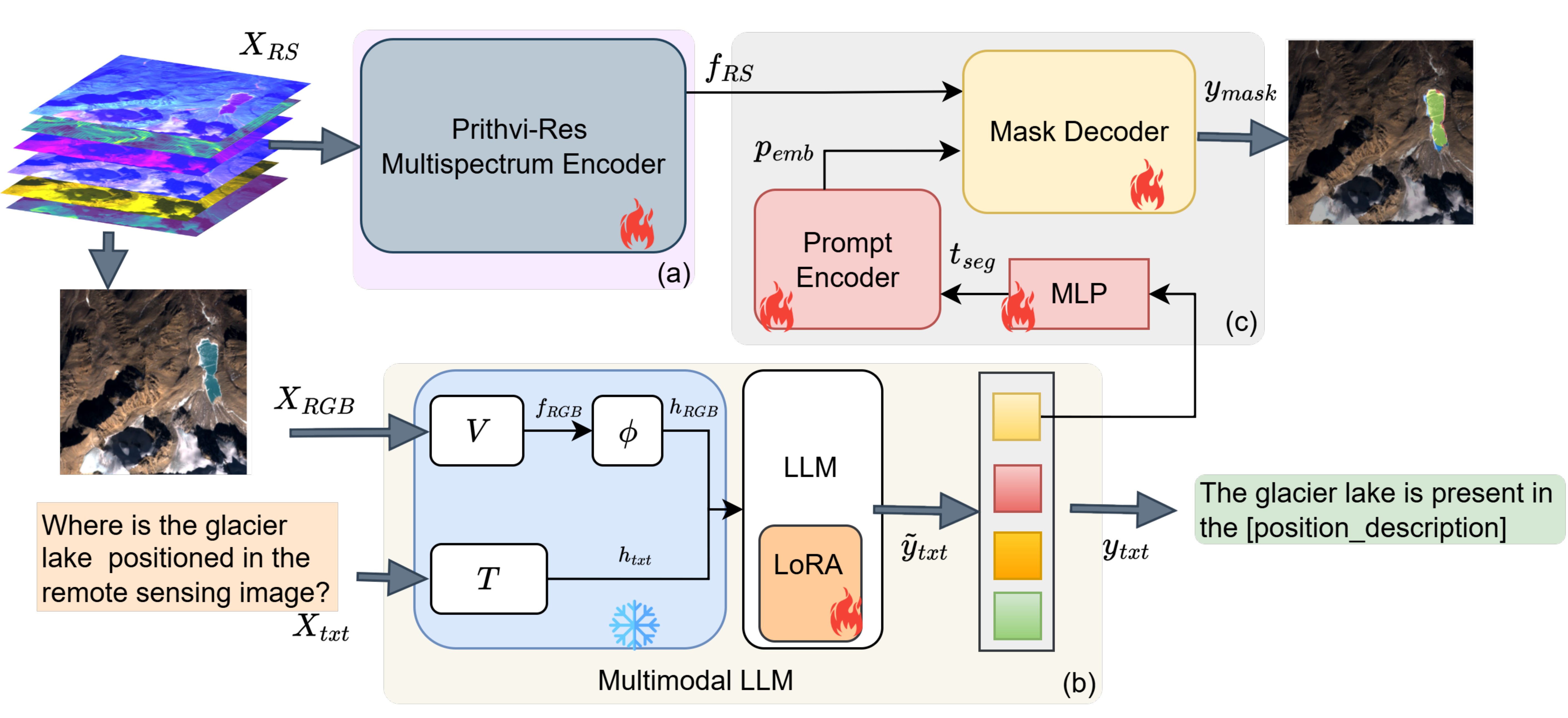}
    \caption{Overview of the proposed architecture for glacial lake segmentation. (a) Prithvi-Res encoder fuses multispectral local and global context, while (b) multimodal LLM generates segmentation-specific tokens from RGB imagery and text. (c) Prompt Mask Decoder aligns these tokens with multispectral features to produce precise spatial masks.}
    \label{fig:block}
\end{figure*}
The proposed architecture (Fig. \ref{fig:block}) integrates multispectral feature extraction with multimodal reasoning to enable precise glacial lake segmentation.  The Prithvi-Res encoder combines a ResNet-34 stem adapted for six-channel input with selected transformer layers from Prithvi-EO v2, fusing local textures and global context into compact features  (Fig. \ref{fig:block} (a)).  In parallel, a multimodal LLM processes RGB imagery and textual instructions, generating positional segmentation-specific tokens adapted with LoRA (Fig. \ref{fig:block} (b)). These tokens condition the Prompt Mask Decoder, which aligns semantic prompts with multispectral features through cross-attention, self-attention, and upsampling (Fig. \ref{fig:block} (c)). Joint optimization with segmentation and text generation losses ensures accurate spatial masks while preserving semantic reasoning. The details are explain in following subsection.

\subsection{Prithvi-Res Multispectral Encoder}

This work presents Prithvi-Res, a hybrid multi-scale feature extraction backbone that integrates a convolutional ResNet stem with hierarchical transformer representations from the Prithvi GFM \cite{szwarcman2024prithvi}. The encoder performs cross-resolution fusion, attention-guided feature propagation, and multi-level refinement to generate a compact yet semantically enriched representation suitable to extract pixel-level features from remote sensing image to support conditional segmentation. We utilized Prithvi-EO-v2 with 100 million parameters, which is based on the lightweight architecture of Prithvi GFM \cite{szwarcman2024prithvi}. This design enables efficient image encoding while reducing computational resource requirements. Prithvi-Res is designed to process six-channel input images and generate a compact latent representation that integrates both local and global contextual information. Formally, the input remote senisnig image is denoted as ${X_{RS} \in \mathbb{R}^{B \times 6 \times H \times W}}$, where $B$ is the batch size, $H$ and $W$ are spatial dimensions, and the six channels correspond to multi-spectral input modalities. At its core, the encoder consists of two parallel branches.  The first branch is a ResNet-based CNN stem, following a ResNet-34 \cite{he2016deep} like configuration adapted for six-channel input images. This produces multi-scale convolutional features

\begin{equation}
\mathrm{ResNetStem6}(x) \rightarrow \{c_2, c_3, c_4, c_5\}
\end{equation}
with each feature map
$ c_i \in \mathbb{R}^{B \times C_i \times \tfrac{H}{s_i} \times \tfrac{W}{s_i}}$ where, $s_i \in \{4,8,16,32\}$, and $C_i \in \{64,128,256,512\}$. These features capture localized spatial patterns, textures, and edge information at progressively coarser resolutions, providing complementary information to the transformer backbone. The Prithvi Transformer backbone models long-range dependencies. Concurrently, Prithvi processes $X_{RS}$ to produce hidden states from transformer layers:

\begin{equation}
    \mathcal{L} = \{l_1, l_2, \dots, l_n\}
\end{equation}

For selected layers $l\in \{2, 5, 8, 11\}$, the CLS token is removed and the remaining tokens are reshaped into a two-dimensional feature map:
\begin{equation}
    F_{prithvi} = \mathrm{reshape}( B, D, H_t, W_t)
\end{equation}
where $D$ is the embedding dimension and $H_t = W_t = \sqrt{N}$ ($N$ = patches) . This preserves spatial structure while embedding global context. To align modalities, both transformer and ResNet features are projected into a common channel dimension:
\begin{equation}
\small
    \tilde{F}_{prithvi} = \mathrm{Conv}_{1\times1}(F_{prithvi}),  \quad  F_{Res} = \mathrm{Conv}_{1\times1}(c_{i})
\end{equation}
and ResNet features are interpolated to transformer resolution:
\begin{equation}
     \tilde F_{Res} = \mathrm{Interp}( F_{Res}, H_t, W_t)
\end{equation}
Fusion is then performed by concatenation followed by a reduction convolution:
\begin{equation}
    Z_l = \mathrm{Conv}_{1\times1}(\mathrm{Concat}(\tilde F_{prithvi} , \tilde F_{Res} ))
\end{equation}

Each fused map $Z_l$ is refined through two 
convolution--batch-normalization--ReLU layers, producing the refined representation $\widehat{Z}_l$. The Feature Boost Block enhances features via attention, and  a top–down Feature Pyramid Network (FPN) propagates fusion information to final output feature maps aggregating information across multiple scales. High-resolution features propagate down to preserve fine details.
The encoder thus outputs $f_{RS} $ a compact yet information-rich tensor that integrates local detail and global context for downstream tasks.

\subsection{MultiModal LLM}
Multimodal LLM backbones such as Flamingo \cite{flamingo}, LLaVA \cite{llava}, Qwen-VL \cite{qwen2vl}, and InternVL \cite{internvl} demonstrate strong high-level visual–language reasoning but remain suboptimal for dense prediction tasks (e.g., segmentation), largely due to their limited spatial awareness, positional grounding, and instance-aware reasoning—capabilities essential for glacial lake delineation. Despite these limitations, their final layer embeddings capture rich semantic information \cite{wen2025tinyvla,liu2024robomamba,trinh2025prs}. Motivated by these findings, we introduce a unified multimodal framework for glacial lake segmentation that directly leverages the native joint embeddings of Multimodal-LLMs. These embeddings are used to condition a masked segmentation decoder that enables both instance-aware and position-aware reasoning, thereby preserving the semantic expressiveness of the backbone while substantially improving spatial understanding in diverse remote-sensing environments.

The remote sensing input ${X_{RS} \in \mathbb{R}^{B \times 6 \times H \times W}}$ccontains six-band multispectral satellite imagery, from which the complete spectral information is processed by the Prithvi-Res encoder to extract features $f_\text{RS}$ for segmentation. In parallel, only the visible RGB bands image, $X_{\text{RGB}} \in \mathbb{R}^{b \times 3 \times w \times h}$, is supplied to the multimodal LLM.  Following a LLaVA-style architecture \cite{llava}, the RGB image first passes through a global visual encoder  $V$, implemented using the CLIP ViT-H/14 model, to extract visual features $f_{RGB}=V(X_{RGB})$. A projection layer $\phi$ then maps these features into the LLM input space, yielding visual tokens $h_{RGB} = \phi(f_{RGB})$. At the same time, textual instructions describing glacial lake positioning are tokenized by the Mistral-7B tokenizer $T$, resulting in text tokens $h_{txt} = T(X_{txt})$ . The multimodal LLM integrates both visual and textual tokens to generate reasoning outputs  descibed as:
\begin{equation}
    \tilde y_{\text{txt}} = F_{\text{LLM}}(h_{\text{RGB}}, h_{\text{txt}})
\end{equation}
where $F_{\text{LLM}}$ parametric function the processing procedure of Mistral \cite{jiang2023mistral7b}. When segmentation masks are required, the output sequence $\tilde y_{\text{txt}}$ includes special segmentation tokens, $t_{seg}$, which are extracted from the final embedding layer of the LLM. These tokens are processed through an $MLP$ to form segmentation prompts that guide the downstream mask decoder with positional reasoning. 
\begin{equation}
    t_{seg} = MLP(\tilde y_{\text{txt}})
\end{equation}
To efficiently adapt the LLM to the glacial lake dataset without incurring the cost of full fine-tuning, the framework employs Low-Rank Adaptation (LoRA) \cite{hu2022lora}, enabling position-aware reasoning while maintaining semantically rich embeddings that guide the downstream mask decoder to produce accurate glacial lake segmentation.

\subsection{Prompt Mask Decoder}

This module generates segmentation masks for glacial lakes using two key inputs: the multispectral feature $f_{\text{RS}}$ from the Prithvi-Res encoder, and the joint image-text embedding  $ p_{emb} = PE( t_{\text{seg}})$, where the prompt features produced by the multimodal LLM are processed through a prompt encoder  ($PE(.)$ ). 
Two projection layers, $\mathcal{F}^{\text{proj}}_{\theta_1}$ and $\mathcal{F}^{\text{proj}}_{\theta_2}$, map both input features into a shared latent space of dimension 256. This enables effective fusion through multi-head cross-attention, aligning glacial lake spatial patterns with textual reasoning. In this process, prompt tokens selectively highlight relevant image regions, serving as text-guided feature modulation. 
\begin{equation}
\small
    A_{\text{fused}} = \text{MHA}\left( \sigma\left( 
    \mathcal{F}^{\text{proj}}_{\theta_1}(f_{\text{RS}}) 
    \mathcal{F}^{\text{proj}}_{\theta_2}(p_{\text{emb}})^{T}
    \right) / \sqrt{d_k} \right)
    \mathcal{F}^{\text{proj}}_{\theta_2}(p_{\text{emb}}),
\end{equation}
Here, $d_k$ is the scaling factor, $\text{MHA}(\cdot)$ denotes multi-head attention, and $\sigma(\cdot)$ represents the softmax function. A residual connection followed by LayerNorm stabilizes training, improves gradient flow, and prevents representational collapse.
To further refine contextual reasoning, the decoder applies self-attention across attended tokens, enabling global dependencies and improving feature coherence. A feed-forward network introduces non-linear transformations that enhance expressiveness and reduce noise. The refined token sequence is reshaped into a spatial feature map and fused element-wise with the original image features, preserving fine-grained boundaries and textures. Finally, a sequence of transposed 2D convolutional layers, each followed by Batch Normalization and ReLU activation, progressively upsamples the fused representation to the original input resolution. The result is the final glacial lake segmentation mask  $y_{\text{mask}}$. 
\subsection{Loss Function}
The training objective is defined as a weighted sum of segmentation and text generation losses, enabling joint optimization of spatial mask prediction and multimodal reasoning. Formally, 
\begin{equation}
    L = \lambda_{\text{seg}} L_{\text{seg}} + \lambda_{\text{txt}} L_{\text{text}},
\end{equation}
where $\lambda_{\text{seg}}$ and $\lambda_{\text{txt}}$ control the relative importance of each component.  The segmentation loss  $L_\text{seg}$ integrates Binary Cross-Entropy (BCE) with Dice loss, a widely adopted strategy in remote sensing segmentation tasks to balance pixel-level accuracy with overlap-based region consistency.  The text generation loss $L_\text{text}$ is implemented as a categorical cross-entropy loss applied to the predicted token logits, ensuring accurate sequence generation aligned with ground truth instructions. This design enables simultaneous learning of spatial segmentation and semantic reasoning, with gradients from both tasks reinforcing each other.


\section{Experiments}
\textbf{Dataset and Metrics.} To evaluate the performance of our model, we used glacial lake datasets from two regions: the Himalayas and the European Alps. The Himalayan dataset, provided by \cite{kaushik2022automated}, originally consists of images of size 300×300×10 with ten channels (B, G, R, NIR, SWIR, TIR, SAR coherence, slope, elevation, and NDWI). Following the findings of \cite{kaushik2022automated}, we selected the six most important channels (B, G, R, NIR, slope, and elevation) as training data.  
For the European Alps, we generated a similar dataset consisting of image–label pairs. Sentinel-2 images from 2020 during the ablation season (July–September) were used, and labels were derived from the global lake dataset provided by \cite{zhang2024heterogeneous}, where pixel value 1 represents lake pixels and 0 represents non-lake pixels. We employed standard metrics to assess segmentation performance. Mean Intersection over Union (mIoU) was used to measure the exact overlap between predicted masks and ground truth, while the Dice score was used to compute similarity by evaluating the harmonic mean of precision and recall.  

\noindent\textbf{Implementation Details.} The dataset contains a total of 1000 images, split into training and testing sets with an 80:20 ratio.  The weights of the text generation loss ($\lambda_{txt}$) and the mask loss ($\lambda_{seg}$) were both set to 1.0. Training was performed with a batch size of 4 using the AdamW optimizer, with a learning rate of $10^{-4}$ and weight decay of $10^{-6}$. The model was trained on an NVIDIA A100 GPU with 40 GB memory.  
For comparison, we considered geospatial foundation models including Prithvi-EO-v2.0 -100/600/300 \cite{szwarcman2024prithvi}, DOFA \cite{xiong2024neural}, UViT (a hybrid of Prithvi and CNN) \cite{jiang2025glacial}, TransNorm (ViT-based) \cite{azad2022transnorm}, and U-Net (CNN-based) \cite{ronneberger2015u}. In addition, we compared GLACIA against reasoning-driven segmentation models such as LISA \cite{lisa} with 7B and 13B versions, as well as PixelLM \cite{ren2024pixellm}.

\section{Results and Discussion}
\subsection{Quantitative Results}
\begin{table*}[!h]
\caption{Comparison of several state-of-the-art supervised methods for glacial lake segmentation. In each column, the highest score is shown in \textbf{bold} and the second-highest score is \uline{underlined}. }
\centering
\label{tab:seg_results}
\resizebox{0.7\linewidth}{!}{%
\begin{tabular}{@{}l|l|llll|llll@{}}
\toprule
\multirow{2}{*}{Methods} & \multirow{2}{*}{Band} & \multicolumn{4}{c|}{Single Glacial Lake} & \multicolumn{4}{c}{Multiple Glacial Lake} \\ \cmidrule(l){3-10} 
                        &                       & IoU      & mIoU     & Dice    & mDice   & IoU      & mIoU     & Dice     & mDice    \\ \cmidrule(r){1-10}

Prithvi 100 \cite{szwarcman2024prithvi}  & B-G-R-NIR-S-E  
& 54.38 & 76.17 & \uline{70.46} & \uline{84.70} & 70.99 & 84.96 & 83.04 & 91.25 \\

Prithvi 300 \cite{szwarcman2024prithvi}  & B-G-R-NIR-S-E  
& 41.84 & 69.62 & 58.99 & 78.84 & 73.33 & 86.19 & 84.61 & 92.07 \\

Prithvi 600 \cite{szwarcman2024prithvi}  & B-G-R-NIR-S-E  
& 53.24 & 75.59 & 69.51 & 84.22 & \uline{75.10} & \uline{87.10} & \uline{85.78} & \uline{92.66} \\

DOFA \cite{xiong2024neural}              & B-G-R-NIR-S-E  
& 28.53 & 60.13 & 44.39 & 70.04 & 55.01 & 76.37 & 70.98 & 84.92 \\

UNet \cite{ronneberger2015u}             & B-G-R-NIR-S-E  
& 18.14 & 58.03 & 21.85 & 60.39 & 58.28 & 78.55 & 67.91 & 83.65 \\

UNet \cite{ronneberger2015u}             & R-G-B          
& 48.47 & 73.41 & 57.17 & 78.16 & 59.30 & 79.01 & 69.75 & 84.55 \\

TransNorm \cite{azad2022transnorm}       & B-G-R-NIR-S-E  
& \uline{56.73} & \uline{77.32} & 67.05 & 82.92 & 64.51 & 81.75 & 75.49 & 87.49 \\

TransNorm \cite{azad2022transnorm}       & R-G-B          
& 39.10 & 68.25 & 47.71 & 73.12 & 40.34 & 69.27 & 50.01 & 74.54 \\

UViT \cite{jiang2025glacial}             & B-G-R-NIR-S-E  
& 47.89 & 72.86 & 58.03 & 78.40 & 59.33 & 79.11 & 69.27 & 84.36 \\ \midrule

GLACIA                                     & B-G-R-NIR-S-E  
& \textbf{69.38} & \textbf{84.01} & \textbf{81.92} & \textbf{90.62} & \textbf{75.70} & \textbf{87.30} & \textbf{86.17} & \textbf{92.81} \\ \bottomrule
\end{tabular}
}
\end{table*}

\textbf{Segmentation Performance}. Our proposed GLACIA model outperforms all other deep learning approaches, including GFMs (Prithvi and DOFA), ViT-based model (TransNorm), the hybrid model (UViT), and U-Net—in both single-lake and multi-lake scenarios (Table \ref{tab:seg_results}). In the case of a single glacial lake, our proposed GLACIA model significantly outperforms all comparative models (Table \ref{tab:seg_results}), exceeding the next best-performing model (TransNorm) by at least 12.65\%. This demonstrates GLACIA’s superior ability to reason over spatial relationships and leverage multi-spectral features. In comparison, traditional architectures such as U-Net using the full spectral input (B-G-R-NIR-S-E) show limited performance (IoU = 18.14, Dice = 21.85), indicating difficulty in leveraging complex spectral and topographic information. Interestingly, U-Net trained only on RGB bands performs much better (IoU = 48.47), suggesting that additional topographic inputs (slope and elevation) may introduce noise or that the limited multi-spectral dataset size reduces convergence efficiency. Among pretrained foundation models, Prithvi 600 provides balanced performance (IoU = 53.24, Dice = 69.51), as expected from a model trained with approximately 600M parameters and designed to generalize well across remote-sensing tasks. GLACIA significantly surpasses all baselines (IoU = 69.38, Dice = 81,92), demonstrating its superior capability in reasoning over spatial relationships and multi-spectral features.In the multiple glacial lake setting, all models generally perform better (Table \ref{tab:seg_results}, likely due to the larger dataset that supports more stable convergence and provides stronger contextual cues. GLACIA further incorporates Instance-Aware Spatial Reasoning, which leverages both lake counts and positional cues to enhance spatial consistency, reduce false detections, and improve discrimination between true and false-positive regions. While Prithvi 600 and TransNorm remain competitive, GLACIA again achieves the highest scores (IoU = 75.70, Dice = 86.17), confirming its robustness in handling positional complexity across diverse multi-lake configurations. 
\begin{table}[!h]
\caption{Comparison with previous related works for reasoning based segmentation}
\label{tab:llm_seg}
\small
\resizebox{1\linewidth}{!}{%
\begin{tabular}{@{}l|l|ll|ll@{}}
\toprule
\multirow{2}{*}{Method} & \multirow{2}{*}{Band} & \multicolumn{2}{l|}{Single Glacial Lake} & \multicolumn{2}{l}{Multiple Glacial Lake} \\ \cmidrule{3-6}
                 &                & mIoU   & mDice  & mIoU   & mDice  \\ \cmidrule{1-6}
LISA 7B \cite{lisa}        & R-G-B          & 57.82 & 61.23 & 60.12 & 67.16 \\
LISA 13B \cite{lisa}       & R-G-B          & 71.33 & 80.40 & 75.66 & 84.26 \\
PixelLM \cite{ren2024pixellm} & R-G-B       & 69.12 & 81.27 & 74.32 & 84.01 \\ \midrule
GLACIA                    & B-G-R-NIR-S-E  & \textbf{84.01} & \textbf{90.62} & \textbf{87.30} & \textbf{92.81} \\
\bottomrule
\end{tabular}
}
\end{table}

Comparative evaluation with reasoning segmentation methods for glacial lake segmentation across single and multiple lake scenarios shows consistent performance by GLACIA (Table \ref{tab:llm_seg}). Models such as LISA 7B, LISA 13B, and PixelLM rely on RGB bands and show moderate success, with LISA 13B achieving the strongest results among them (mIoU = 71.33, mDice = 80.40 for single lakes). PixelLM also performs competitively, particularly in Dice scores. However, GLACIA, which integrates broader spectral bands (B-G-R-NIR-S-E), significantly outperforms all baselines, reaching mIoU = 84.01 and mDice = 90.62 for single lakes, and even higher for multiple lakes, demonstrating superior reasoning segmentation capability.
\begin{table}[!h]
\caption{Performance of reasoning-based segmentation models using language metrics, reflecting accuracy in lake instance and spatial reasoning.}
\label{tab:reason}
\resizebox{\linewidth}{!}{%
\begin{tabular}{@{}lllllll@{}}
\toprule
Method                & ROUGE  & METEOR & BLEU-1 & BLEU-2 & BLEU-3 & BLEU-4 \\ \midrule
LISA 7B \cite{lisa}          & 0.3450 & 0.3002 & 0.2382 & 0.1470 & 0.0631 & 0.0412 \\
LISA 13 B \cite{lisa}       & 0.4332 & 0.3148 & 0.3316 & 0.1969 & 0.0748 & 0.0471 \\
PixelLM \cite{ren2024pixellm} & 0.3520 & 0.3071 & 0.2484 & 0.1506 & 0.0643 & 0.0413 \\ \midrule
GLACIA            & \textbf{0.5606} & \textbf{0.4775} & \textbf{0.4272} & \textbf{0.2623} & \textbf{0.1665} & \textbf{0.1089} \\ \bottomrule
\end{tabular}%
}
\end{table}

\noindent\textbf{Position Reasoning Context Results.} Table \ref{tab:reason} compares the contextual reasoning performance of segmentation models applied to glacial lake analysis. LISA 7B produces relatively weak alignment (ROUGE = 0.3450, METEOR = 0.3002), indicating limited ability to capture semantic consistency between predicted and reference outputs. LISA 13B improves substantially (ROUGE = 0.4332, METEOR = 0.3148), reflecting stronger contextual reasoning and better lexical overlap. PixelLM achieves similar results to LISA 7B, showing moderate contextual fidelity. In contrast, GLACIA achieves the highest scores (ROUGE = 0.5606, METEOR = 0.4775, BLEU-4 = 0.1089), which technically signifies superior precision in generating coherent, multi-token sequences and stronger semantic alignment. These outcomes demonstrate that the proposed approach not only improves positional segmentation but also enhances contextual reasoning, producing outputs that are more faithful to ground truth descriptions and semantically richer than existing multimodal LLM-based baselines.
\begin{figure*}[!ht]
    \centering
    \includegraphics[width=1\linewidth]{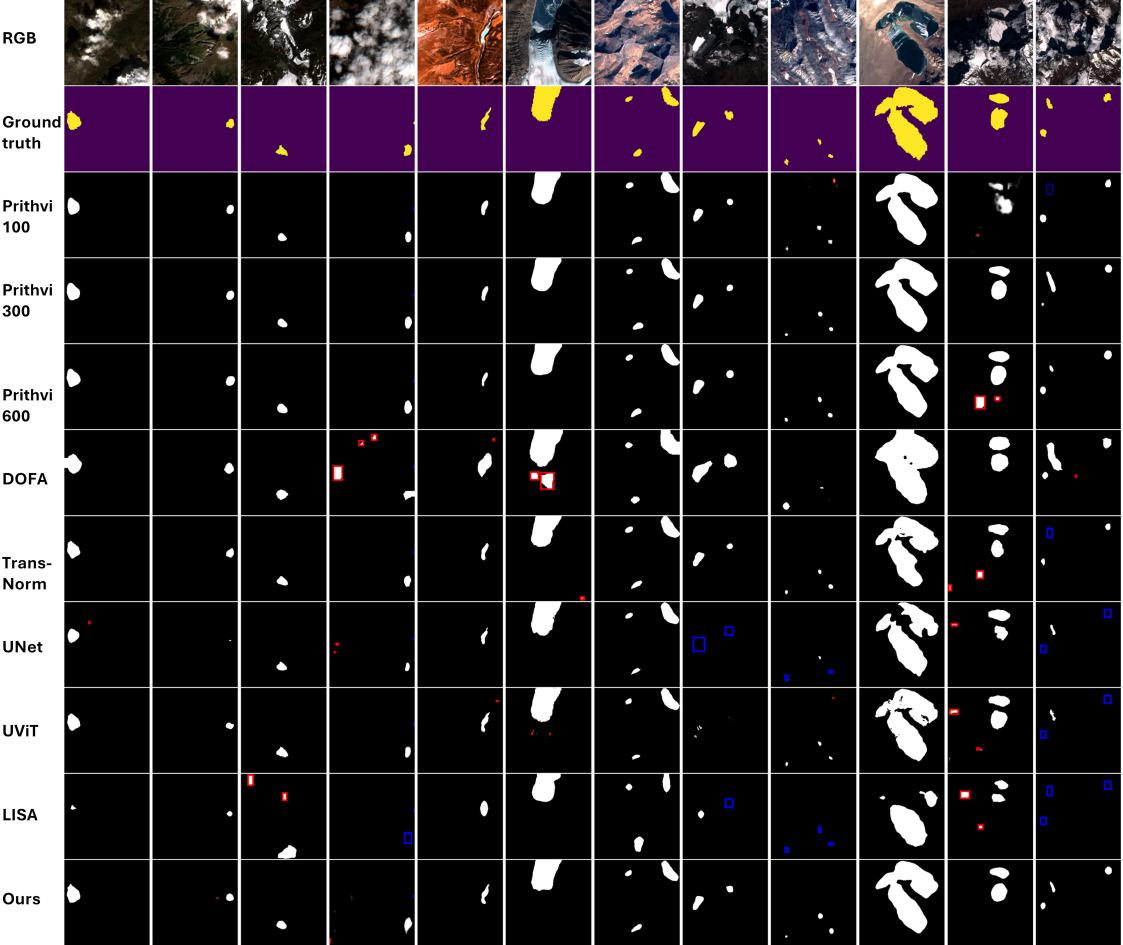}
    \caption{Qualitative comparison of glacial lake segmentation. Red boxes indicate false positives, blue boxes indicate false negatives. Our reasoning-enhanced model accurately captures small and irregular lakes, reducing both errors compared to baseline methods.}
    \label{fig:seg_result}
\end{figure*}
\subsection{Qualitative Analysis}
Our qualitative analysis reaffirms the stable and superior performance of the proposed GLACIA compared to other deep learning models (Fig. \ref{fig:seg_result}). GLACIA successfully segments glacial lakes of varied sizes and turbidity and, most notably, delivers promising results under challenging conditions such as shadows and cloud cover—where most models, including widely used U-Net, DOFA GFM, and LISA, struggle (Fig. \ref{fig:seg_result}). GLACIA achieves higher fidelity by using instance-aware spatial reasoning, reducing both false positives and false negatives for precise segmentation even in challenging terrains. Minor boundary inaccuracies remain, but overall results highlight GLACIA’s effectiveness.\begin{figure*}[!ht]
    \centering
    \includegraphics[width=1\linewidth]{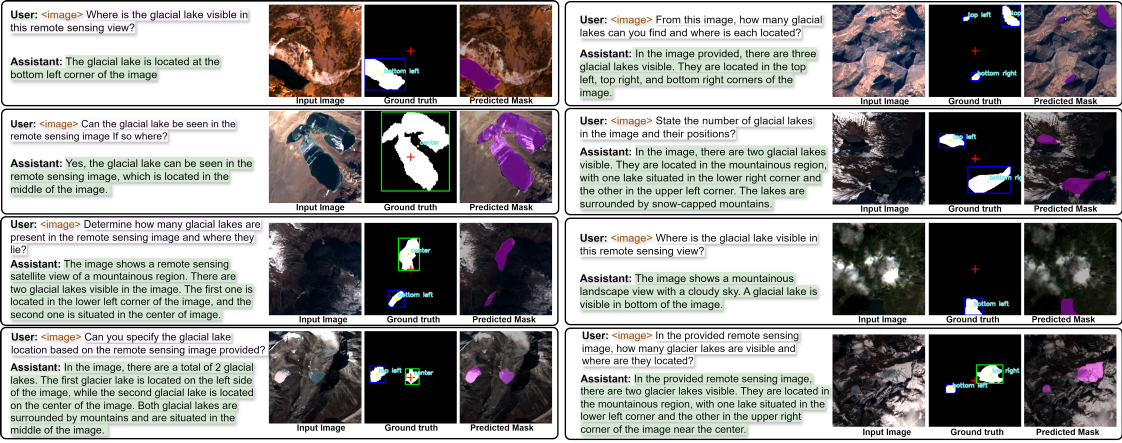}
    \caption{Examples of reasoning outputs and corresponding segmentation masks from test samples used in evaluation.}
    \label{fig:result_text}
\end{figure*}
Fig. \ref{fig:result_text} illustrates how user queries guide segmentation masks, making it easier to interpret how the model’s reasoning aligns with expected visual outcomes. Each example combines a query about lake location or count, the assistant’s reasoning, and the predicted mask. Position-aware reasoning enhances interpretability by linking textual cues with visual outputs. For instance, when asked to identify lakes in specific regions such as the “bottom left corner” or “middle of the image,” the model correctly interprets spatial references and generates masks that closely match ground truth. In queries involving lake counts and positional details, the model shows robustness by detecting multiple lakes and their relative positions, even in complex mountainous or cloudy scenes. This reasoning segmentation synergy addresses limitations of traditional models, which often misinterpret ambiguous regions. While minor deviations occur under dense cloud cover or overlapping terrain, results highlight the promise of multimodal approaches for accurate, context-aware glacial lake detection. \begin{table}[!h]
\caption{Ablation study comparing different encoder variants}
\centering
\label{tab:ablation}
\resizebox{0.85\linewidth}{!}{%
\begin{tabular}{@{}llllll@{}}
\toprule
Backbone               & Parameters & IoU    & mIoU   & Dice    & mDice  \\
\midrule
Prithvi 100   (Frozen) & 100M       & 56.77 & 77.50 & 72.42 & 85.77 \\
Prithvi 100   (FT)     & 100M       & 68.24 & 83.47 & 81.12 & 90.23 \\
Prithvi 300   (Frozen) & 300M       & 66.08 & 82.37 & 79.58 & 89.45 \\
Prithvi 300   (FT)     & 300M       & 73.42 & 86.18 & 84.67 & 92.07 \\ \midrule
Prithvi-Res (FT)        & 137.14M    & 75.70 & 87.30 & 86.17 & 92.81 \\
\bottomrule
\end{tabular}%
}
\end{table}


\subsection{Ablation Study}
Table \ref{tab:ablation} presents an ablation study on different backbone variants for glacial lake segmentation, comparing frozen and fully fine-tuned (FT) settings. Frozen backbones, such as Prithvi 100 (IoU = 56.77, mDice = 85.77), provide moderate performance, showing the utility of pretrained features alone. Full fine-tuning allows the model to adapt to lake-specific spatial patterns, improving results significantly (e.g., Prithvi 300 FT: IoU = 73.42, mDice = 92.07). Our Prithvi-Res (FT) backbone, which integrates multi-scale pretrained features with full tuning, achieves the best performance (IoU = 75.70, mDice = 92.81), highlighting the effectiveness of hybrid feature fusion for reasoning-aware segmentation.

\subsection{Discussion}
\textbf{Advantages.} The comparative evaluation highlights clear distinctions between GLACIA and conventional CNN- and ViT-based segmentation models. CNN models excel at extracting local pixel-level features, while ViT models enhance global context modeling. However, both approaches remain limited to producing segmentation masks without explicit reasoning about positional attributes. As a result, their outputs lack interpretability and require expert post-analysis to translate technical results into actionable insights. In contrast, GLACIA unifies segmentation with natural language reasoning through multimodal LLMs, producing instance-aware outputs that delineate lakes with higher accuracy and describe their positions, counts, and scene layout in human-interpretable terms. Such capability bridges the gap between technical segmentation and practical decision-making, offering disaster managers and policymakers a more intuitive understanding of glacial environments. By integrating visual and linguistic modalities, GLACIA transforms segmentation from a purely pixel-level task into a reasoning-driven process, establishing a new paradigm for interpretable remote sensing applications.

\noindent\textbf{Limitation.} Despite these advantages, GLACIA faces certain limitations in producing precise segmentation masks and consistent textual descriptions. Segmentation challenges are most evident in cases involving small lakes (0.002–0.5 km²), cloudy conditions, and frozen lakes. A key limitation in textual description arises when handling multiple lake scenarios. CLIP-based vision encoders used in multimodal LLMs typically operate on reduced-resolution feature maps, which can cause small or thin lakes to be missed or merged, degrading accurate instance separation. Moreover, descriptions of lake “positions” are generated at the language level and may become inconsistent or even hallucinatory when many visually similar lakes are present. These issues are further compounded by hallucination tendencies in the LLaVA model.

\noindent\textbf{Future Outlook.} Future improvements should focus on enhancing reasoning with environmental context by incorporating geospatial metadata (e.g., latitude, longitude, elevation), multi-temporal imagery to capture lake dynamics, and synthetic aperture radar (SAR) data to improve performance under cloudy conditions.

\section{Conclusion}
We introduce GLACIA, the first framework to unify glacial lake segmentation with instance-aware positional reasoning. Our approach integrates a lightweight multispectral Prithvi-Res encoder with a vision–language model, enabling intuitive and conversational interaction for segmentation tasks enriched with contextual instance awareness. To support this capability, we introduce the GLake-Pos dataset pipeline, generating spatially grounded question–answer pairs for reasoning-driven segmentation. Overall, our proposed GLACIA outperforms state-of-the-art methods such as U-Net, Prithvi, DOFA, TransNorm, and UViT. In addition, unlike previous approaches, GLACIA provides natural language descriptions of segmentation outputs, offering interpretable insights that can have direct implications for simple yet effective disaster preparedness and decision-making.
{   
    \small
    \bibliographystyle{ieeetr}
    \bibliography{ref}
}

\cleardoublepage
\newpage
\appendix
\section{Appendix}
\addcontentsline{toc}{section}{Supplementary Material} 
\renewcommand{\thesection}{\Alph{section}} 
\renewcommand{\thetable}{\thesection.\arabic{table}} 
\renewcommand{\thefigure}{\thesection.\arabic{figure}} 
\subsection{Generation of GLake-Pos: Question-Answer Pairs}
To construct the reasoning-enhanced supervision used in GLACIA, we generate spatially grounded natural-language explanations directly from the segmentation masks. For each image, connected-component analysis is applied to the binary glacier-lake mask to identify individual lakes and extract their geometric attributes, including bounding box, centroid, and distance from the image center. Based on these measurements, each lake is assigned a positional label—such as top left, bottom right, or center—together with a proximity descriptor (near or far from the center). These spatial descriptors are then inserted into multiple linguistic templates to produce diverse reasoning statements.
\begin{figure}[h]
    \centering
    \includegraphics[width=1\linewidth]{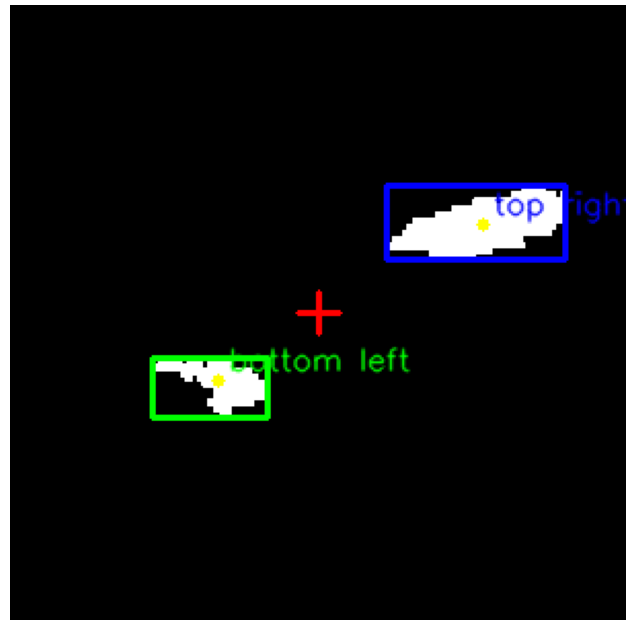}
    \caption{The position of Glacial Lake form the center }
    \label{fig:pos}
\end{figure}

For example, in the image shown in Fig. \ref{fig:pos}, the mask contains two distinct glacial lakes. The first lake is located in the top right, far from the center of the image, while the second lake appears in the bottom left, near the center. Using our template-based generation system, GLACIA produces a corresponding reasoning-aware annotation such as: “There are two glacial lakes: the 1st is in the top right, far from the center, and the 2nd is in the bottom left, near the center.” Such structured reasoning annotations are stored alongside each image and question, forming a comprehensive dataset that teaches the model not only to recognize glacial lakes but also to articulate their spatial relationships.
 The process consists of three major stages:
(i) extraction of glacial lake positions from binary masks,
(ii) creation of structured descriptive sentences, and
(iii) filling question--answer templates to produce natural-language
annotations. The overall pipeline is described below.

\textbf{(1) Determining the Spatial Position of Each Glacial Lake}

For every mask image, we performed connected-component analysis using
\texttt{cv2.connectedComponentsWithStats}.  
Each connected component corresponds to a single glacial lake.  
For each detected lake, we computed:
\begin{itemize}
    \item its bounding box,
    \item the centroid location $(x_c, y_c)$,
    \item its relative position with respect to the image center,
    \item and proximity to the center (``near the center'' or ``far from the center'').
\end{itemize}

Each lake is then assigned one of the following spatial categories:
\textit{top left}, \textit{top right}, \textit{bottom left},
\textit{bottom right}, or \textit{center}, with additional proximity
information attached.  
For example, a resulting descriptor may be:
\textit{``bottom right, near the center''}.

\textbf{(2) Constructing Descriptive Answer Sentences}\\
Let $N$ denote the total number of detected glacial lakes.
For each lake $i$, we generate an ordinal label
(e.g., ``1st'', ``2nd'', ``3rd'') and pair it with its computed
positional description.

Depending on $N$, three levels of answer construction are used:
\begin{itemize}
    \item For $N=1$, a single-sentence answer template is chosen from a set
    of 10 stylistic alternatives.
    \item For $N=2$, a dual-lake template is used, explicitly describing the
    locations of the 1st and 2nd lakes.
    \item For $N\geq 3$, the answer consists of:
    \begin{enumerate}
        \item a high-level count sentence (e.g., 
        ``There are 5 glacial lakes in the image.''), and
        \item individual per-lake sentences generated randomly from a set
        of per-lake templates such as:  
        ``The \{number\_position\} glacial lake lies in the
        \{position\_description\}.''  
    \end{enumerate}
\end{itemize}

All per-lake sentences are concatenated to form the final answer string.

\textbf{(3) Template-Based Question Generation}\\
A large collection of question templates, written in JSON format, was used
to generate natural-language questions.  
Each template requests the number of glacial lakes and their locations.
Examples include:
\begin{itemize}
    \item ``In the provided remote sensing image, how many glacial lakes are visible and where are they located?''
    \item ``Identify all glacial lakes in the image and give their respective positions.''
    \item ``Provide the glacial lake count and describe the location of each glacial lake.''
\end{itemize}

For each image, one template is sampled at random.  
Placeholders such as \texttt{\{all\_glacial\_sentences\}} and
\texttt{\{total\_number\}} are dynamically replaced based on the detected
lake count and generated descriptions.

\textbf{(4) Final Q--A Output}\\
Each image thus receives:
\begin{itemize}
    \item a natural-language question asking for the number and location of the glacial lakes, and
    \item a systematically constructed answer describing each lake in ordinal order with precise spatial reasoning.
\end{itemize}

An example of final answer for three lakes might be:
\begin{quote}
    ``There are 3 glacial lakes in the image.  
    The 1st glacial lake lies in the top left, near the center.  
    The 2nd glacial lake can be seen at bottom right, far from the center.  
    The 3rd glacial lake appears in the top right, near the center.''
\end{quote}

\subsection{Position-Based Question--Answer Templates.}
The following templates are used for generating spatial reasoning
questions and answers, where \texttt{\{position\_description\}} is
automatically derived from the mask-based positional analysis described above:

\begin{itemize}
    \item \textbf{Q:} In the provided remote sensing image, where is the glacial lake located?\\
          \textbf{A:} The glacial lake can be observed in the \{position\_description\} area of the image.

    \item \textbf{Q:} Specify the glacial lake location in this remote sensing view.\\
          \textbf{A:} The glacial lake is clearly seen in the \{position\_description\} region.

    \item \textbf{Q:} Can you detect where the glacial lake is in this remote sensing image?\\
          \textbf{A:} The glacial lake is present in the \{position\_description\} area.

    \item \textbf{Q:} Which area of the remote sensing image shows the glacial lake?\\
          \textbf{A:} The glacial lake is seen in the \{position\_description\}.

    \item \textbf{Q:} From this remote sensing image, what is the glacial lake position?\\
          \textbf{A:} The glacial lake is located in the \{position\_description\} part.

    \item \textbf{Q:} Could you point out where the glacial lake is located in this remote sensing scan?\\
          \textbf{A:} The glacial lake can be observed in the \{position\_description\} area of the scan.

    \item \textbf{Q:} What part of the image contains the glacial lake in this remote sensing?\\
          \textbf{A:} In this remote sensing image, the glacial lake is found in the \{position\_description\} section.

    \item \textbf{Q:} Identify the region in this remote sensing image that shows the glacial lake.\\
          \textbf{A:} The region showing the glacial lake is the \{position\_description\}.

    \item \textbf{Q:} Can you specify the glacial lake location based on the remote sensing image provided?\\
          \textbf{A:} Based on the image, the glacial lake lies in the \{position\_description\} region.

    \item \textbf{Q:} From this remote sensing image, where would you say the glacial lake is located?\\
          \textbf{A:} Judging from the image, the glacial lake is located at the \{position\_description\}.
          
    \item \textbf{Q:} In the provided remote sensing image, can you locate the glacial lake?\\
          \textbf{A:} The glacial lake appears in the \{position\_description\} area.

    \item \textbf{Q:} Which section of this remote sensing image contains the glacial lake?\\
          \textbf{A:} The glacial lake is present in the \{position\_description\} section.

    \item \textbf{Q:} Where does the glacial lake appear in the remote sensing image?\\
          \textbf{A:} The glacial lake is situated in the \{position\_description\} region.

    \item \textbf{Q:} Highlight the glacial lake location in this remote sensing scan.\\
          \textbf{A:} The glacial lake is located in the \{position\_description\} part of the image.

    \item \textbf{Q:} Can the glacial lake be seen in the remote sensing image? If so, where?\\
          \textbf{A:} The glacial lake can be observed in the \{position\_description\}.

    \item \textbf{Q:} In the remote sensing image provided, what is the glacial lake position?\\
          \textbf{A:} The glacial lake is found in the \{position\_description\} region.

    \item \textbf{Q:} Could you identify the glacial lake in this remote sensing image?\\
          \textbf{A:} The glacial lake is located in the \{position\_description\} area.

    \item \textbf{Q:} Where is the glacial lake visible in this remote sensing view?\\
          \textbf{A:} The glacial lake can be observed in the \{position\_description\} region.

    \item \textbf{Q:} Which part of the remote sensing scan shows the glacial lake?\\
          \textbf{A:} The glacial lake is present in the \{position\_description\} section.

    \item \textbf{Q:} Based on this remote sensing image, where is the glacial lake located?\\
          \textbf{A:} The glacial lake lies in the \{position\_description\} area.

    \item \textbf{Q:} Identify where the glacial lake is in the remote sensing image.\\
          \textbf{A:} The glacial lake is found in the \{position\_description\} section.

    \item \textbf{Q:} Can you determine the glacial lake location in this remote sensing image?\\
          \textbf{A:} The glacial lake is located in the \{position\_description\} region.

    \item \textbf{Q:} Point out the glacial lake in the remote sensing image.\\
          \textbf{A:} The glacial lake can be observed in the \{position\_description\} part.

    \item \textbf{Q:} In the provided remote sensing scan, which section shows the glacial lake?\\
          \textbf{A:} The glacial lake appears in the \{position\_description\} area.

    \item \textbf{Q:} Where is the glacial lake situated in this remote sensing image?\\
          \textbf{A:} The glacial lake is seen in the \{position\_description\} region.

    \item \textbf{Q:} Specify the area of this remote sensing image that contains the glacial lake.\\
          \textbf{A:} The glacial lake is located in the \{position\_description\} section.

    \item \textbf{Q:} Which part of the remote sensing view shows the glacial lake?\\
          \textbf{A:} The glacial lake is present in the \{position\_description\} area.

    \item \textbf{Q:} Can the glacial lake be identified in this remote sensing scan?\\
          \textbf{A:} Yes, the glacial lake is in the \{position\_description\} region.

    \item \textbf{Q:} Highlight the glacial lake location in the provided remote sensing image.\\
          \textbf{A:} The glacial lake is situated in the \{position\_description\} area.

    \item \textbf{Q:} In this remote sensing image, can you specify the glacial lake region?\\
          \textbf{A:} The glacial lake is found in the \{position\_description\}.

    \item \textbf{Q:} Where does the glacial lake appear in the remote sensing scan?\\
          \textbf{A:} The glacial lake is observed in the \{position\_description\} part.

    \item \textbf{Q:} Point out where the glacial lake is located in this remote sensing image.\\
          \textbf{A:} The glacial lake can be seen in the \{position\_description\} section.

    \item \textbf{Q:} Which region of the remote sensing image shows the glacial lake?\\
          \textbf{A:} The glacial lake is visible in the \{position\_description\} region.

    \item \textbf{Q:} Based on this remote sensing view, identify the glacial lake location.\\
          \textbf{A:} The glacial lake is in the \{position\_description\} area.

    \item \textbf{Q:} Can you locate the glacial lake in the provided remote sensing scan?\\
          \textbf{A:} The glacial lake appears in the \{position\_description\} section.

\end{itemize}

\subsection{Instance-Aware Position Question--Answer Templates.}

The Q\&A generation for glacial lake segmentation is based on the number and positions of detected lakes in an image. We use different templates depending on whether there is a single lake, exactly two lakes, or more than two lakes.

\begin{description}
    \item[Single Glacial Lake:] If the image contains only one glacial lake, the question is paired with an answer template such as:
    \begin{itemize}
        \item \textbf{Question:} How many glacial lakes are present in this image?  
        \textbf{Answer:} There is one glacial lake located in \texttt{\{position\_description\}}.
        \item \textbf{Question:} Count the number of glacial lakes.  
        \textbf{Answer:} A single glacial lake can be seen at \texttt{\{position\_description\}}.
        \item \textbf{Question:} Identify the locations of glacial lakes.  
        \textbf{Answer:} The only glacial lake in this image lies in the \texttt{\{position\_description\}}.
    \end{itemize}

    \item[Two Glacial Lakes:] If the image contains exactly two glacial lakes, the templates include:
    \begin{itemize}
        \item \textbf{Question:} How many glacial lakes are present in this image?  
        \textbf{Answer:} There are two glacial lakes: the 1st is in \texttt{\{position\_description[0]\}}, and the 2nd is in \texttt{\{position\_description[1]\}}.
        \item \textbf{Question:} Count and locate all glacial lakes.  
        \textbf{Answer:} Two glacial lakes can be seen here: the 1st lies in \texttt{\{position\_description[0]\}} and the 2nd is in \texttt{\{position\_description[1]\}}.
        \item \textbf{Question:} How many glacial lakes and where are they positioned?  
        \textbf{Answer:} There are exactly two glacial lakes: the first is at \texttt{\{position\_description[0]\}} and the second at \texttt{\{position\_description[1]\}}.
    \end{itemize}

    \item[Multiple Glacial Lakes:] If the image contains more than two glacial lakes, we use a combination of total count and per-lake sentences:
    \begin{itemize}
        \item \textbf{Question:} How many glacial lakes can be observed?  
        \textbf{Answer:} There are \texttt{\{total\_number\}} glacial lakes. \texttt{\{all\_glacial\_sentences\}}
        \item \textbf{Question:} Provide the count and locations of glacial lakes.  
        \textbf{Answer:} glacial lake count in this scan is \texttt{\{total\_number\}}. \texttt{\{all\_glacial\_sentences\}}
        \item \textbf{Question:} Describe the positions of glacial lakes in the image.  
        \textbf{Answer:} Observed glacial lakes: \texttt{\{total\_number\}}. \texttt{\{all\_glacial\_sentences\}}
    \end{itemize}
\end{description}

\subsubsection{Notes on Placeholders}
\begin{itemize}
    \item \texttt{\{position\_description\}}: Position of a single glacial lake relative to the image center (e.g., \textit{top left, near the center}).
    \item \texttt{\{position\_description[0]\}}, \texttt{\{position\_description[1]\}}: Positions of the first and second glacial lakes when exactly two are present.
    \item \texttt{\{total\_number\}}: Total number of glacial lakes in the image.
    \item \texttt{\{all\_glacial\_sentences\}}: Dynamically generated per-lake description for images with more than two glacial lakes.
\end{itemize}

\end{document}